\documentclass[pdflatex,sn-mathphys-num]{sn-jnl}

\usepackage{graphicx}%
\usepackage{multirow}%
\usepackage{amsmath,amssymb,amsfonts}%
\usepackage{amsthm}%
\usepackage{mathrsfs}%
\usepackage[title]{appendix}%
\usepackage{xcolor}%
\usepackage{textcomp}%
\usepackage{manyfoot}%
\usepackage{booktabs}%
\usepackage{algorithm}%
\usepackage{algorithmicx}%
\usepackage{algpseudocode}%
\usepackage{listings}%

\theoremstyle{thmstyleone}%
\theoremstyle{thmstyletwo}%

\theoremstyle{thmstylethree}%

\begin{document}

\title[Self-Supervised Z-Slice Augmentation for 3D Bio-Imaging via Knowledge Distillation]{Self-Supervised Z-Slice Augmentation for 3D Bio-Imaging via Knowledge Distillation}

\author[1]{\fnm{Alessandro} \sur{Pasqui}}
\author[1]{\fnm{Sajjad} \sur{Mahdavi}}
\author[2]{\fnm{Benoit} \sur{Vianay}}
\author[2]{\fnm{Alexandra} \sur{Colin}}
\author[3]{\fnm{Rémi} \sur{Dumollard}}
\author[3]{\fnm{Alex} \sur{McDougall}}
\author[4]{\fnm{Yekaterina} \sur{A. Miroshnikova}}
\author[5,6,7]{\fnm{Elsa} \sur{Labrune}}
\author*[1]{\fnm{Hervé} \sur{Turlier}}\email{herve.turlier@college-de-france.fr}

\affil[1]{Center for Interdisciplinary Research in Biology (CIRB), Collège de France, Université PSL, CNRS, INSERM, 75005 Paris, France}
\affil[2]{Université Grenoble-Alpes, CEA, CNRS, INRA, Interdisciplinary Research Institute of Grenoble, CytoMorpho Lab, Grenoble, France}
\affil[3]{Laboratory of Developmental Biology of the Villefranche-sur-Mer, Institute of Villefranche-sur-Mer, Sorbonne University, CNRS, Villefranche-sur-Mer, France}
\affil[4]{Laboratory of Molecular Biology, National Institute of Diabetes and Digestive and Kidney Diseases, National Institutes of Health, Bethesda, MD, 20892, USA}
\affil[5]{Department of Reproductive Medicine, Hospices Civils de Lyon, Bron, France}
\affil[6]{INSERM U1208 Stem Cells and Brain Institute, Bron, France}
\affil[7]{Claude Bernard Lyon 1 University, Lyon, France}

\abstract{Three-dimensional biological microscopy has significantly advanced our understanding of complex biological structures. However, limitations due to microscopy techniques, sample properties or phototoxicity often result in poor z-resolution, hindering accurate cellular measurements. Here, we introduce ZAugNet, a fast, accurate, and self-supervised deep learning method for enhancing z-resolution in biological images. By performing nonlinear interpolation between consecutive slices, ZAugNet effectively doubles resolution with each iteration. Compared on several microscopy modalities and biological objects, it outperforms competing methods on most metrics. Our method leverages a generative adversarial network (GAN) architecture combined with knowledge distillation to maximize prediction speed without compromising accuracy. We also developed ZAugNet+, an extended version enabling continuous interpolation at arbitrary distances, making it particularly useful for datasets with nonuniform slice spacing. Both ZAugNet and ZAugNet+ provide high-performance, scalable z-slice augmentation solutions for large-scale 3D imaging. They are available as open-source frameworks in PyTorch, with an intuitive Colab notebook interface for easy access by the scientific community.}




\maketitle

\section{Introduction}\label{sec1}

Recent advances in three-dimensional biological imaging have profoundly impacted the life sciences, enabling detailed volumetric visualization of complex biological structures. High-precision optical sectioning techniques \cite{Zhang2025}, including confocal and light-sheet microscopy \cite{Huisken2004,Tomer2012,Chen2014}, structured illumination microscopy \cite{Gustafsson2000, Heintzmann2009}, and super-resolution microscopy \cite{Betzig2006,Rust2006,Mortensen2010}, now allow researchers to explore fine structural details, such as cellular organelles, tissue architectures, and developing organisms, which are essential for understanding fundamental biological processes.

However, a persistent challenge in this field is the relatively lower resolution along the axial (z) dimension compared to the lateral (x-y) dimensions. This limitation arises from inherent physical, technical or biological constraints linked to imaging systems. Balancing between imaging speed, spatial resolution, and light exposure is crucial to mitigate issues such as photo-bleaching and photo-toxicity in samples, as well as to accommodate the required imaging depth, temporal resolution, and the duration of the experiment. These constraints necessitate trade-offs that impact the quality and resolution of the acquired images \cite{Pawley2006, Scherf2015}. Most 3D biological and biomedical imaging techniques suffer from anisotropic resolution, caused by axial elongation of the optical point spread function (PSF) and low axial sampling rates in fast volumetric acquisitions, which compromises accurate measurements of cellular properties, such as volume or shape. As computational tools become increasingly integral to microscopy, there is a growing need for scalable, data-driven solutions that enhance z-resolution while preserving fine biological details.

To address this challenge, relatively few computational methods have been explored, and the literature on z-slice augmentation for three-dimensional images remains sparse. Common approaches include bi-linear interpolation and the more accurate yet computationally intensive bi-cubic interpolation. However, these methods exhibit significant limitations in both the accuracy and quality of interpolated images. Deep learning algorithms have driven significant breakthroughs in biological imaging in recent years. Applications of these methods include fluorescence signal prediction from label-free images \cite{Ounkomol2018, Christiansen2018}, restoration of low-SNR fluorescent images \cite{Weigert2018, Buchholz2019, Sreehari2017}, image segmentation \cite{Stringer2021, Schmidt2018}, single-molecule localization in super-resolution microscopy \cite{Nehme2018, Ouyang2018}, and improving microscopy image resolution in both lateral and axial dimensions \cite{deHaan2019, Wang2019, Qiao2021}. In 2022, He et al. \cite{He2022} introduced Super-Focus, a convolutional neural network (CNN) based method to predict images interpolating between two consecutive focal planes for three-dimensional biological images. Using a generative adversarial network (GAN) architecture, Super-Focus directly generates intermediate images from two consecutive focal planes. The method was developed for \textit{in vitro} fertilization (IVF) imaging, where transmitted light microscopy in time-lapse human embryo recordings captures only a limited number of focal planes per time point. In 2024, Priessner et al. \cite{Priessner2024} introduced a deep learning-based approach inspired by advances in computer vision, specifically video frame interpolation, which increases frame rates to produce slow-motion sequences. Their method reinterprets the temporal axis of videos as the z-axis in 3D microscopy. They adapted two content-aware frame interpolation (CAFI) networks, originally designed for video frame prediction: Zooming SlowMo \cite{Xiang2020, Xiang2021} and Depth-Aware Video Frame Interpolation \cite{Bao2019}. These models, optimized for predicting intermediate frames in video sequences, demonstrated potential for enhancing both spatial and temporal resolutions of biological image stacks and time-lapses post-acquisition. Despite the potential of their approach, the algorithms reused by Priessner et al. exhibit notable limitations in both accuracy and computational efficiency, resulting in time-consuming training and prediction phases. The CAFI methods rely on complex neural network architectures with a high number of parameters (11.1 million for Zooming SlowMo and 24.0 million for Depth-Aware Video Frame Interpolation). While these models perform effectively in video frame interpolation with large training datasets, they may be impractical for biological applications, where datasets are typically limited and highly complex. The high computational demand and large model sizes pose challenges in training these models accurately on such specialized data. Additionally, these limitations hinder real-world applicability, as processing large-scale datasets is often essential for obtaining statistically significant insights into biological processes. Furthermore, while Priessner et al. have made their source code available via a Colab notebook and integrated it into the ZeroCostDL4Mic platform, its maintenance appears to present challenges. Currently, the implementation is affected by deprecated packages and compatibility issues, making it difficult to run without modifications. The outdated implementation combined with the limited availability of the datasets used in their study poses constraints on benchmarking their method against newer or alternative approaches.

To overcome these challenges, we introduce ZAugNet, a self-supervised deep learning method tailored for accurate and rapid interpolation of contiguous focal planes in post-acquisition 3D biological datasets. ZAugNet leverages a GAN architecture to reconstruct high-resolution stacks with superior accuracy while integrating a knowledge distillation scheme in its generator, which ensures the model remains lightweight during prediction, significantly improving computational efficiency and mitigating slow inference times.

\begin{figure}[htbp]
    \centering
    \includegraphics[width=1.\textwidth]{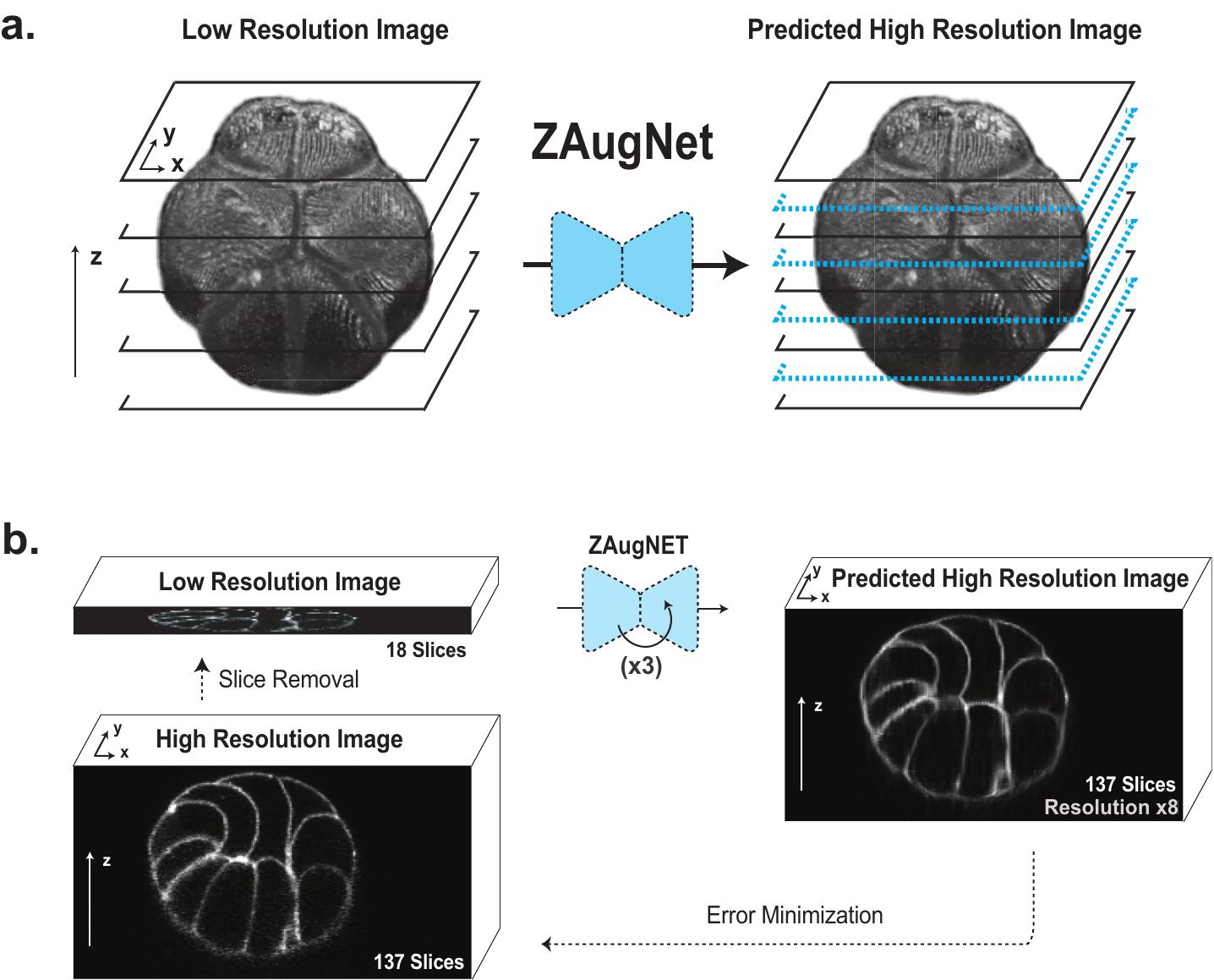}
    \caption{\textbf{ZAugNet principle.} \textbf{a)} Graphical representation of ZAugNet’s prediction process, demonstrating its ability to double the axial resolution of a 3D image (here a 16-cell \textit{Phallusia Mammillata} ascidian embryo). \textbf{b)} Self-supervised training scheme of ZAugNet: on the left, selected slices are removed from the stack to serve as ground truth. On the right, ZAugNet predicts the missing slices, optimizing its parameters by minimizing the error between the predicted slices and the original focal planes prior to their removal.}\label{fig:figure1}
\end{figure}

To assess its performance, ZAugNet was benchmarked against existing classical and state-of-the-art interpolation techniques, including bicubic interpolation and deep learning approaches, such as Super-Focus and CAFI. This assessment was conducted across diverse biological imaging datasets spanning multiple microscopy domains. A comprehensive evaluation using quantitative metrics—including Fréchet Inception Distance (FID), root mean square error (RMSE), peak signal-to-noise ratio (PSNR), and structural similarity index measure (SSIM)—along with qualitative assessments and time performance analyses, consistently demonstrated that ZAugNet outperforms existing methods in terms of interpolation accuracy, training efficiency, and prediction speed. Furthermore, ZAugNet was extended into ZAugNet+, a novel approach enabling continuous interpolation at arbitrary relative distances between input slices. This enhancement greatly expands the method’s applicability, allowing it to handle nonuniform slice spacing and making it particularly suited for datasets with variable z-resolution.

ZAugNet is introduced as a ready-to-use, open-source framework implemented in PyTorch, with the code freely accessible on GitHub. To enhance usability, a dedicated Colab notebook with an intuitive interface has been developed, making it more accessible to the broader scientific community. This notebook, along with the corresponding datasets and pretrained models, enables researchers to train new ZAugNet models and perform predictions efficiently. 

\section{Results}\label{sec2}

\subsection{Model and architecture}

ZAugNet is a deep learning-based method designed for fast and accurate interpolation of contiguous focal planes in 3D biological imaging. With each iteration, ZAugNet doubles the axial resolution of an initial low-resolution image, progressively enhancing it until the desired high resolution along the z-axis is achieved (Figure \ref{fig:figure1}a).

As a self-supervised method, ZAugNet eliminates the need for annotated data. It starts with a small set of high-resolution 3D images acquired from a microscope. To generate training data, the method removes every second slice, creating a low-resolution dataset with half the original resolution. Using two consecutive slices from this low-resolution dataset, ZAugNet predicts the missing focal plane by minimizing the error relative to the corresponding original triplet in the high-resolution dataset (Figure \ref{fig:figure1}b).

\begin{figure}[htbp]
    \centering
    \includegraphics[width=1.\textwidth]{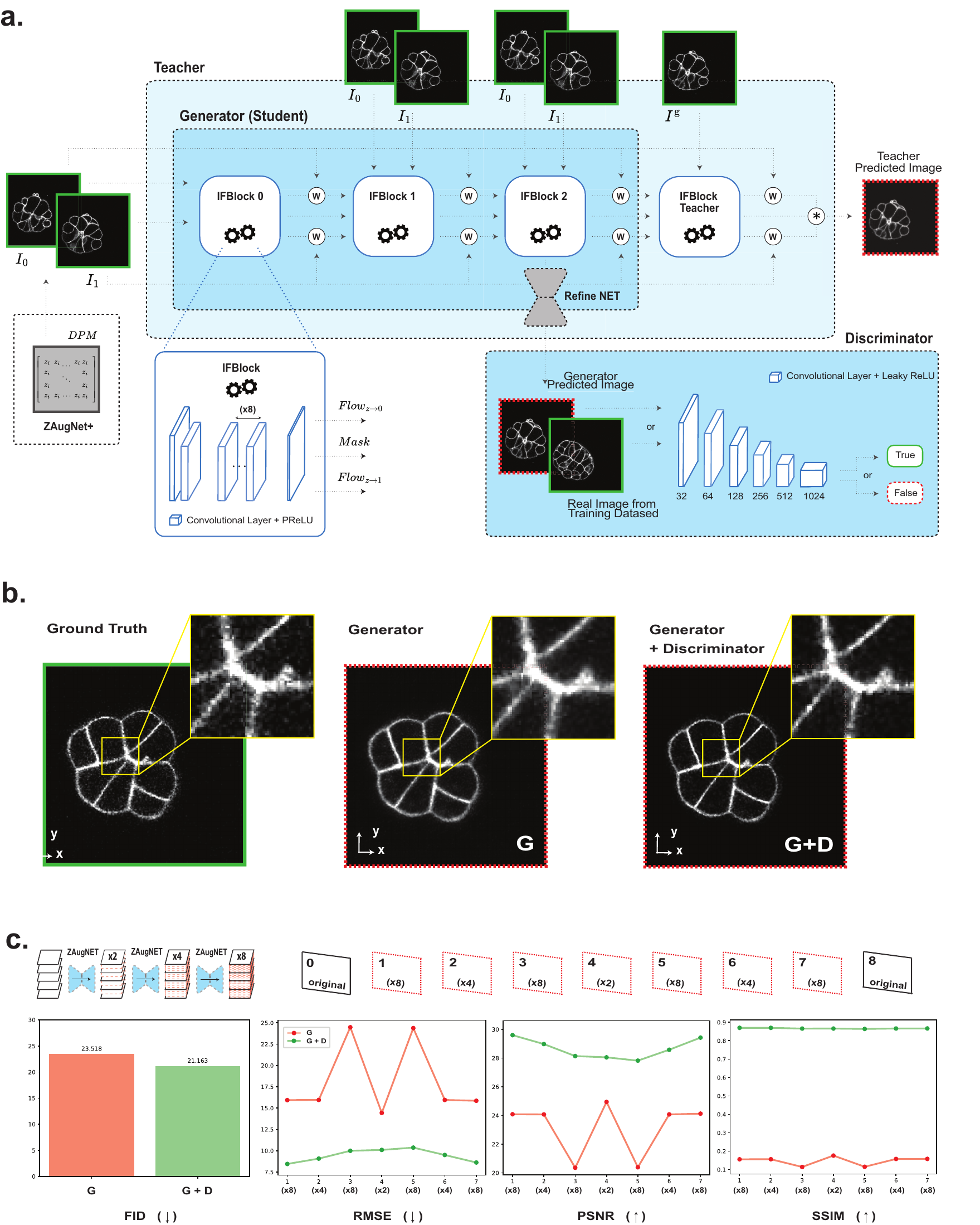}
    \caption{\textbf{Neural network architecture and evaluation.} \textbf{a)} ZAugNet’s architecture: it follows a GAN framework, where the Generator and Discriminator interact to improve the accuracy and reliability of generated biological images. Through a knowledge distillation process, the Teacher network transfers information to the Generator (Student), enabling a more efficient and lightweight design for predictions. ZAugNet+ extends this approach by allowing continuous interpolation between two slices, with users specifying the relative position of the interpolated slice using a Digital Propagation Matrix (DPM) that defines the relative distance $z_i$ to the target plane. 
    \textbf{b)} Direct vs. Adversarial Training in ZAugNet: visual comparison of interpolation results using Generator-only (direct training) versus adversarial training with both the Generator and Discriminator, shown alongside the ground truth (16-cell \textit{Phallusia Mammillata} ascidian embryo).
    \textbf{c)} Quantitative evaluation using Fréchet Inception Distance (FID), where lower values indicate greater similarity to the ground truth. Additional metrics, including RMSE, PSNR, and SSIM, assess the inter-stack average error to further compare the effectiveness of both training approaches.}\label{fig:figure2}
\end{figure}

To non-linearly interpolate between two contiguous focal planes, ZAugNet utilizes a Generative Adversarial Network (GAN) architecture, consisting of a Generator and a Discriminator. Specifically, it employs a Wasserstein GAN with Gradient Penalty (WGAN-GP) to address common training challenges, such as mode collapse and unstable convergence dynamics \cite{NIPS2017_892c3b1c}. The Generator is inspired by the well-established work of Huang et al. \cite{huang2022rife} in video frame interpolation and incorporates convolutional blocks that efficiently estimate intermediate optical flows in an end-to-end manner. Additionally, our approach integrates a knowledge distillation framework \cite{RevModPhys.65.499, hinton2015distilling}, where ZAugNet’s Generator (Student) is a smaller, optimized subset of a larger Teacher network. The Teacher is trained in a self-supervised manner, as described earlier, and its knowledge is transferred to the Student without loss of validity. The lighter and specialized Student network is then used for predictions, ensuring that during inference, ZAugNet’s Generator remains lightweight, significantly improving prediction speed. The Discriminator is based on the well-known work of Gulrajani et al. \cite{NIPS2017_892c3b1c} and operates antagonistically with the Generator to refine output quality. Specifically, the Discriminator is trained to distinguish between real and generated images, while the Generator simultaneously learns to produce increasingly realistic outputs to deceive the Discriminator. Our model maintains a limited complexity, with 10.6 million parameters for the Generator and 11.19 million parameters for the Discriminator, ensuring feasibility for training on small biological datasets while enabling lightweight and efficient inference. See Methods for further details on the architecture (Figure \ref{fig:figure2}a). \\

\subsection{Direct versus adversarial training in ZAugNet}

ZAugNet was initially trained on a dataset of 139 high-resolution 3D image stacks, each containing 36 z-slices, depicting an ascidian embryo at various developmental stages, acquired via fluorescence microscopy with membrane staining (see Methods). Two separate training configurations were performed using identical hyperparameters: one with the Generator alone and another incorporating adversarial training with both the Generator and Discriminator. The trained ZAugNet models were then applied three consecutive times on 127 low-resolution 3D images, each containing 18 slices, to achieve the target z-resolution of 137 slices, representing an 8-fold resolution increase. ZAugNet doubles the initial number of slices $n$ minus one in each iteration, the final interpolated stack has $2n - 1$, that is one slice fewer than a perfect doubling due to the inability to interpolate the last slice in the stack, which lacks a paired slice for interpolation.

The impact of incorporating the Discriminator into ZAugNet’s architecture is demonstrated in Figure \ref{fig:figure2}b. Using only the Generator produces smooth interpolations, which may be suitable for tasks such as preprocessing before segmentation, but are less effective when preserving image texture is crucial. By integrating the Discriminator, the model generates more realistic interpolations, maintaining the texture and noise of the fluorescent dye around the sample. This improvement is quantified using the Fréchet Inception Distance (FID), which assesses the quality of generated images by comparing their distribution to that of the ground truth. The lowest FID value, indicating the highest image fidelity, was achieved with the GAN-based model, where both the Generator and Discriminator were trained together.

Additional evaluation metrics—Root Mean Squared Error (RMSE), Peak Signal-to-Noise Ratio (PSNR), and Structural Similarity Index (SSIM)—were used to measure the average error across the seven generated slices between each pair of original images, after applying ZAugNet three consecutive times to the low-resolution stacks. Across all metrics, the average inter-stack error was consistently lower for images generated by ZAugNet trained with adversarial learning, demonstrating the advantages of training both the Generator and Discriminator together (Figure \ref{fig:figure2}c).

\subsection{Comparison with state-of-the-art interpolation methods}

ZAugNet’s performance was compared against other image interpolation methods, including classical bicubic interpolation and the machine learning-based CAFI \cite{Priessner2024}. Initially, the CAFI implementation provided in the Colab notebook on the ZeroCostDL4Mic platform was non-functional due to deprecated dependencies and compatibility issues. To enable its use, we downloaded the Depth-Aware Video Frame Interpolation (DAIN) implementation locally, thoroughly reviewed and updated the code to restore full functionality for training and predictions on the analyzed datasets.

\begin{figure}[htbp]
\centering
\includegraphics[width=1.\textwidth]{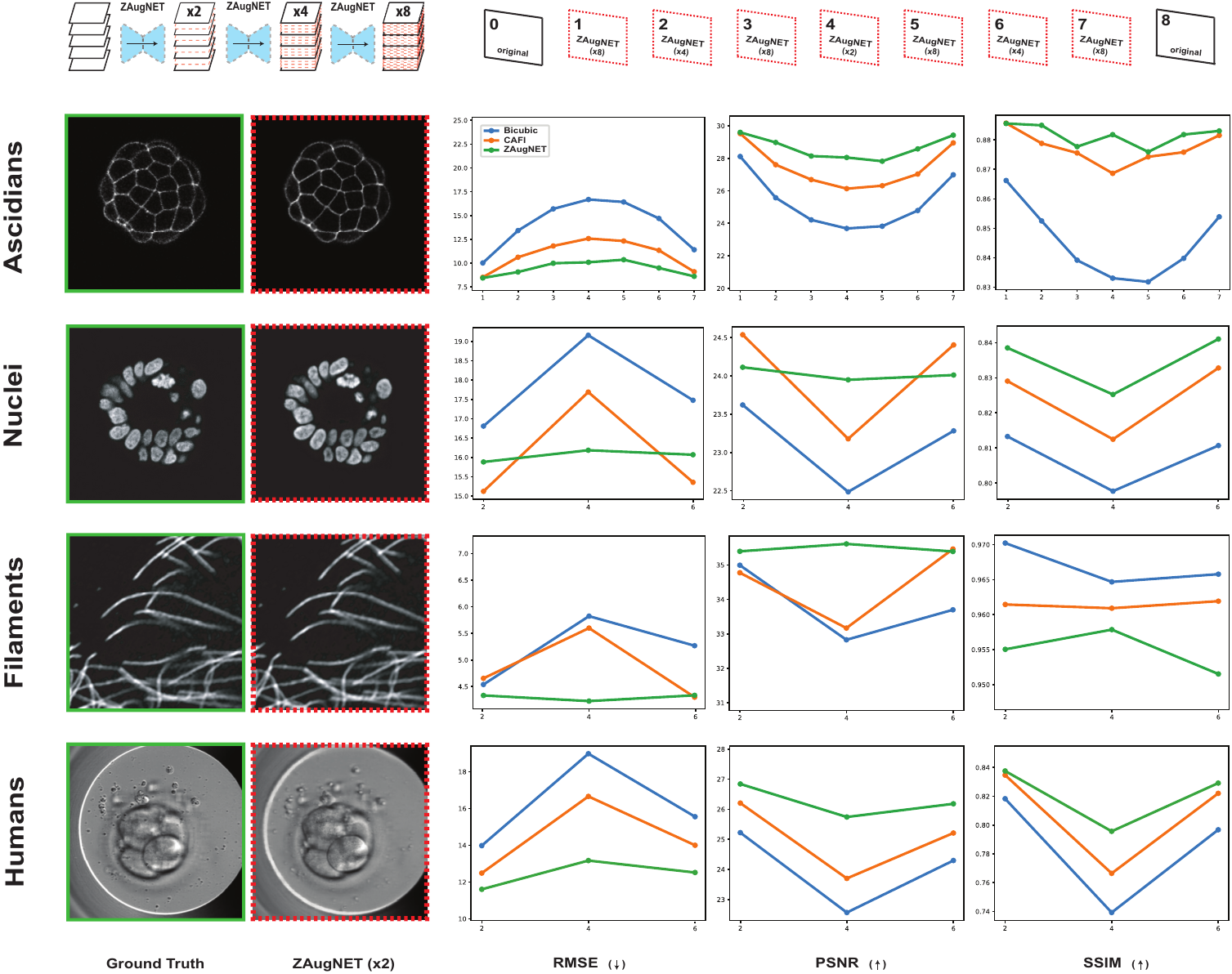}
\caption{\textbf{Comparison of ZAugNet, CAFI, and Bicubic Interpolation Methods.} Each method was applied to four different datasets to achieve the same high-resolution as the available ground truth stacks, resulting in an 8-fold axial resolution increase for ascidian embryos and a 4-fold increase for cell nuclei, filaments of microtubules, and human embryos. The inter-stack average error was quantified using RMSE, PSNR, and SSIM metrics. On the left, the first column presents ground truth slices, while the second column shows the corresponding ZAugNet-predicted slices obtained from the first interpolation step on the low-resolution stacks (2-fold increase). These slices represent the cases where ZAugNet’s performance is, on average, the lowest across all evaluated metrics.}\label{fig:figure3}
\end{figure}

All methods were first benchmarked in terms of training and prediction time, as well as evaluated using standard computer vision metrics—RMSE, PSNR, and SSIM—to compare their outputs against ground truth images. The benchmarking covered a variety of datasets, featuring samples with diverse shapes, textures, and microscopy techniques, including ascidian embryos, cell nuclei, filaments of microtubules (fluorescence microscopy) and human embryos (transmitted light microscopy). The results, shown in Figure \ref{fig:figure3} and summarized in Table \ref{tab:comparison}, highlight ZAugNet’s superior speed in both training and testing phases, along with higher reliability across all datasets according to each evaluated metric.

ZAugNet was designed to leverage parallel processing by distributing input data across multiple GPUs through batch dimension chunking, enabling both training and prediction to utilize two GPUs in the case presented here. In contrast, the CAFI implementation lacks this feature, restricting computations to a single GPU, which significantly reduces computational speed during both training and inference, as seen in Table \ref{tab:comparison}.

\begin{sidewaystable}[htbp]
\begin{tabular}{@{}l|ccccccc@{}}
\toprule
\multicolumn{8}{c}{\textbf{Benchmarks with state-of-the-art methods}} \\
\multicolumn{8}{c}{} \\ \midrule
\textbf{Dataset} (Resolution)\ \ & 
\textbf{Method} &
\textbf{\# Parameters} & 
\multicolumn{2}{c}{\textbf{Runtime}} & 
\multicolumn{3}{c}{\textbf{Metrics}}  \\ 
{\scriptsize \# Image triplets} & & \textbf{Million} & \textbf{Training} & \textbf{Prediction}
& \textbf{RMSE\ $\downarrow$} & \textbf{PSNR\ $\uparrow$} & 
\textbf{SSIM\ $\uparrow$}  
\\
& & \textbf{\small{(Generator + Discriminator)}} & &
& & & \\
\midrule
\textbf{Ascidians} (8-fold)
& Bicubic    & - & - & - & 16.68 & 23.68 & 0.815  \\
& CAFI   & 24.03 & 40h55 & 24:43 min & 12.58 & 26.13 & 0.852 \\
{\scriptsize Training: 5004} \\ {\scriptsize Inference: 2286} &  ZAugNet  &  21.87 &  2h04 &  8:40 min &  \textbf{10.09} &  \textbf{28.04}  &  \textbf{0.865}\\
&   &  \small{(\ 10.68\ \ +\ \ 11.19\ )} &  &  &  &  & \\
&  ZAugNet+ &  21.88 &  07h04 &  \textbf{7:57 min} &  11.10 &  27.22  &  0.862\\
&   &  \small{(\ 10.69\ \ +\ \ 11.19\ )} &  &  &  &  & \\
\midrule
\textbf{Nuclei} (4-fold)
& Bicubic  & - & - & - & 19.15 & 22.48 & 0.797  \\
& CAFI  & - & 24h05 & 09:27 min & 17.68 & 23.17 & 0.812  \\
{\scriptsize Training: 2981} \\ \scriptsize{Inference: 370} &  ZAugNet  &  - &  01h09 &  \textbf{00:41 min} &  \textbf{16.18} &  \textbf{23.94}  &  \textbf{0.825}\\
&  ZAugNet+  &  - &  04h01 &  01:16 min &  16.92 &  23.55  &  0.823\\
\midrule
\textbf{Filaments} (4-fold)
& Bicubic    & - & - & - & 5.82 & 32.87 & \textbf{0.964}  \\
& CAFI    & - & 48h20 & 17:08 min & 5.59 & 33.16 & 0.960  \\
{\scriptsize Training: 6048} \\ {\scriptsize Inference: 720} &  ZAugNet  &  - &  03h02 &  \textbf{01:14 min} &  \textbf{4.22} &  \textbf{35.60}  &  0.959\\
&  ZAugNet+  &  - &  14h18 &  02:28 min &  5.49 &  33.33  &  0.958\\
\midrule
\textbf{Humans} (4-fold)
& Bicubic    & - & - & - & 18.97 & 22.56 & 0.739  \\
& CAFI    & - &  23h04 & 09:35 min & 16.65 & 23.69 & 0.766  \\
{\scriptsize Training: 4165} \\ {\scriptsize Inference: 2499} &  ZAugNet &  - &  01h42 &  \textbf{02:58 min} &  13.16 &  25.74  &  0.795\\
&  ZAugNet+  &  - &  11h15 &  06:03 min &  \textbf{12.29} &  \textbf{26.33}  &  \textbf{0.806}\\
\bottomrule
\end{tabular}
\vspace{0.25cm}
\caption{\textbf{Performance Comparison of ZAugNet, CAFI, and bicubic interpolation methods.} ZAugNet and CAFI were initialized randomly and trained from scratch on identical training datasets shown in the first column. Despite training, CAFI achieved better prediction accuracy using the pre-trained model on the Vimeo90K-septuplet video dataset \cite{Xue2019} provided by the authors. Consequently, this pre-trained model was used for comparison with ZAugNet. The last three columns present the average error in terms of RMSE, PSNR, and SSIM metrics, corresponding to the predicted slices obtained from the initial application of each algorithm on the low-resolution stacks (2-fold axial resolution increase).}\label{tab:comparison}
\end{sidewaystable}

\subsection{Practical applications of ZAugNet for biological image analysis}

While ZAugNet has demonstrated reliability in interpolating biological images based on common computer vision metrics, its true value lies in its practical utility for biologists seeking to enhance z-resolution. It can be applied to improve visualization during rendering or serve as a preprocessing step before established pipelines, such as segmentation workflows. Therefore, we applied ZAugNet to three different case studies to showcase its effectiveness as a robust preprocessing method for quantitative analysis across different biological samples.

\subsubsection{Cell volumes analysis}
    
\begin{figure}[htbp]
\centering
\includegraphics[width=1.\textwidth]{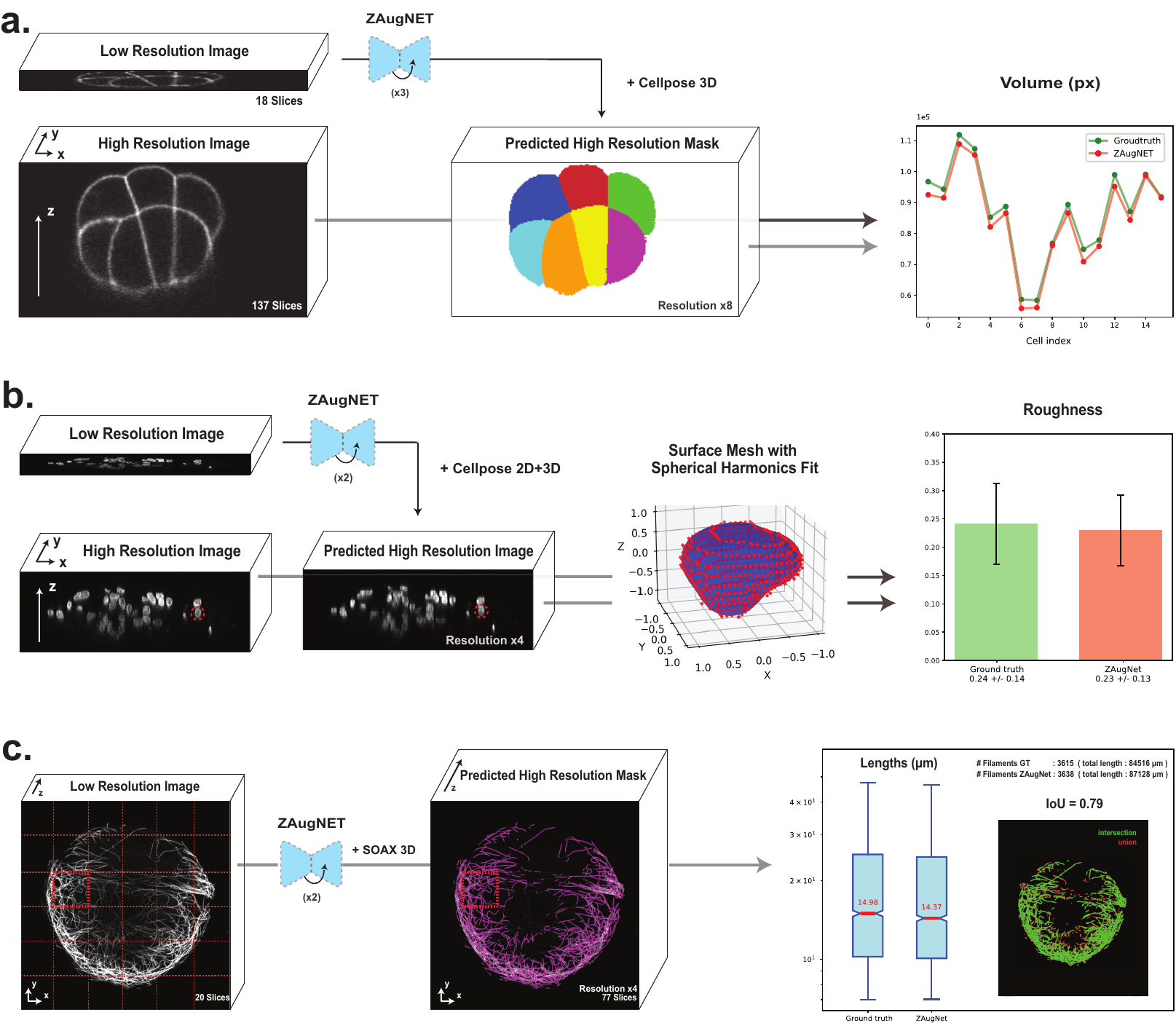}
\caption{\textbf{Benchmarking on microscopy image analysis tasks.} \textbf{a)} Cell volume conservation in ZAugNet-augmented 16-cell stage ascidian embryo images. A low-resolution image stack (18 slices) is axially augmented using ZAugNet, applied iteratively three times to achieve an 8-fold resolution increase (137 slices). Both the ZAugNet-augmented stack and the original high-resolution ground truth are segmented using the 3D Cellpose pipeline. Segmentation masks are aligned, and each cell in the ZAugNet-augmented mask is matched to its corresponding cell in the ground truth mask based on maximum pixel overlap. The plot (right) quantifies cell volume conservation (pixel counts) between the ZAugNet-augmented and ground truth segmentations, demonstrating the accuracy of ZAugNet in preserving biological structures. 
\textbf{b)} Roughness quantification in ZAugNet-augmented cell nuclei images. Four low-resolution image stacks of cell nuclei, each containing a different number of slices but acquired with the same axial resolution, undergo two iterative applications of ZAugNet, resulting in a 4-fold axial resolution increase per stack. Both the ZAugNet-augmented stacks and the original high-resolution ground truth stacks are segmented using a custom pipeline combining 2D and 3D Cellpose. For each nucleus in the segmentation masks, a surface point cloud is extracted by identifying the external pixels of each segmented structure. The point cloud is fitted using spherical harmonics, and their expansion coefficients are used to compute nuclear roughness measurements. The bar plot (right) presents the mean roughness values and their relative standard deviations for both the ground truth and ZAugNet-augmented datasets.
\textbf{c)} Lengths conservation in ZAugNet-augmented microtubule filament images. (Left) Maximum projection of a low-resolution microtubule filaments image (20 slices) and (Center) corresponding higher-resolution image (77 slices), augmented through two successive applications of ZAugNet to achieve a 4-fold resolution increase and segmented into filamentous structures using SOAX 3D. (Right) 36 images obtained after z-resolution augmentation were analyzed using the 3D SOAX pipeline, and results were compared to the ground truth of same resolution. A box plot compares the cumulative filament length between predicted and ground truth images, and the number of detected filaments in indicated at the top. A visual coomparison of the IoU score of 0.79 shows that ZAugNet accurately predicts microtubule structure and spatial continuity.}\label{fig:figure4}
\end{figure}

In Figure \ref{fig:figure4}a, a ZAugNet model trained on 5,004 consecutive triplets of ascidian embryo images was applied three consecutive times to a low-resolution 3D image of a 16-cell stage embryo, originally containing 18 slices, to achieve an 8-fold z-resolution increase. The resulting high-resolution stack of 137 slices was then segmented using the well-established 3D Cellpose pipeline \cite{Stringer2021}. For comparison, the same segmentation pipeline was applied to the ground truth stack, which also contained 137 original slices at the same resolution.

To assess segmentation accuracy, the masks obtained from the ZAugNet-augmented stack and the ground truth stack were compared based on volume overlap for each of the 16 corresponding cells. Each cell in the first mask was mapped to its counterpart in the second mask by identifying the cell with the highest number of overlapping pixels. The volume analysis results, shown in the last column of Figure \ref{fig:figure4}a, demonstrate that cell volumes were conserved across both masks, confirming that ZAugNet produces an accurate output that Cellpose can segment effectively, similar to the ground truth image.

The best-performing ZAugNet model for this analysis was the one trained with the Generator alone, as expected, since Cellpose performs optimally on smoothed, pre-blurred images, where noise is minimized.

\subsubsection{Nuclei roughness analysis}

In Figure \ref{fig:figure4}b, a ZAugNet model was trained on 2,981 consecutive triplets of images depicting cell nuclei and tested twice consecutively on four low-resolution 3D images, achieving a 4-fold z-resolution increase for each image. The primary objective was to quantify nuclear roughness in the ZAugNet-augmented images and compare it with the roughness in the ground truth images.

To ensure accurate preservation of nuclear shape and fine structural details, which are critical for roughness quantification, the ZAugNet model was trained using adversarial learning, incorporating both the Generator and Discriminator. The high-resolution output was segmented using 2D Cellpose, as 3D Cellpose often produces jagged nuclear contours due to image noise, which does not reflect the actual biological structure. However, since 2D segmentation does not retain label consistency along the z-axis, a second segmentation mask was generated using 3D Cellpose to map corresponding nuclei across slices. The 2D segmentation defined the nuclear contours, while the 3D segmentation established correlations along the z-axis. Each 2D label was mapped to the 3D label with the highest pixel overlap, ensuring consistent nucleus labeling across slices.

After segmentation, a 3D point cloud was extracted from the contours of each labeled nucleus. For each nuclear point cloud, the surface radius $ R $ (relative to the geometric center) was reconstructed. To obtain a compact 3D spectral representation of nuclear surface deformations, real spherical harmonic coefficients \cite{mietke2018dynamics} were computed, extending up to an angular number of $ l_{\text{max}} = 5 $. For a rotation-invariant and size-independent characterization of surface roughness, the power spectrum of radial out-of-plane deformations was calculated and normalized by the average shell radius. A final single-valued statistic for surface roughness, $ Ro $, was computed, representing the total power contained in angular numbers $ l \geq 3 $. By excluding long-wavelength modes, $ Ro $  specifically captures fine-scale roughness contributions to nuclear surface deformations (see Methods for details).

Following the same analysis protocol for both the ZAugNet-augmented images and the ground truth images, the roughness analysis results are shown in the last column of Figure \ref{fig:figure4}b. The ZAugNet-augmented measurements align well with the ground truth within the error bars, demonstrating high fidelity in nuclear roughness quantification.

\subsubsection{Filament length analysis}

In Figure \ref{fig:figure4}c, a ZAugNet model trained on 6,048 consecutive triplets of microtubule filament images was applied twice iteratively to 36 low-resolution 3D patches extracted from a larger image, initially containing 20 slices, achieving a 4-fold z-resolution increase per patch. The resulting high-resolution stacks, each comprising 77 slices, were reassembled, and the complete image was analyzed in 3D using the well-established SOAX software \cite{Xu2015}. For comparison, the same pipeline was applied to the ground truth image, which also contained 77 original slices at the same resolution.

To evaluate structural preservation, filament length distributions were quantified and compared between the ZAugNet-augmented and ground truth images. The results, shown in the last column of Figure \ref{fig:figure4}c, include a box plot illustrating the filament length distribution, with median values indicated in red. Additionally, the total number of detected filaments and their cumulative length were computed for both datasets. 

To further assess spatial consistency, binary segmentation masks were generated from the SOAX filament coordinates. These masks underwent post-processing steps, including binary dilation and Gaussian smoothing, before being compared using the Intersection-over-Union (IoU) metric. The final IoU score of 0.79 demonstrates that ZAugNet effectively preserves microtubule filament structures, even in densely entangled regions, maintaining accurate filament lengths and spatial organization.

\subsection{ZAugNet+ for continuous interpolations}

\begin{figure}[htbp]
\centering
\includegraphics[width=1.\textwidth]{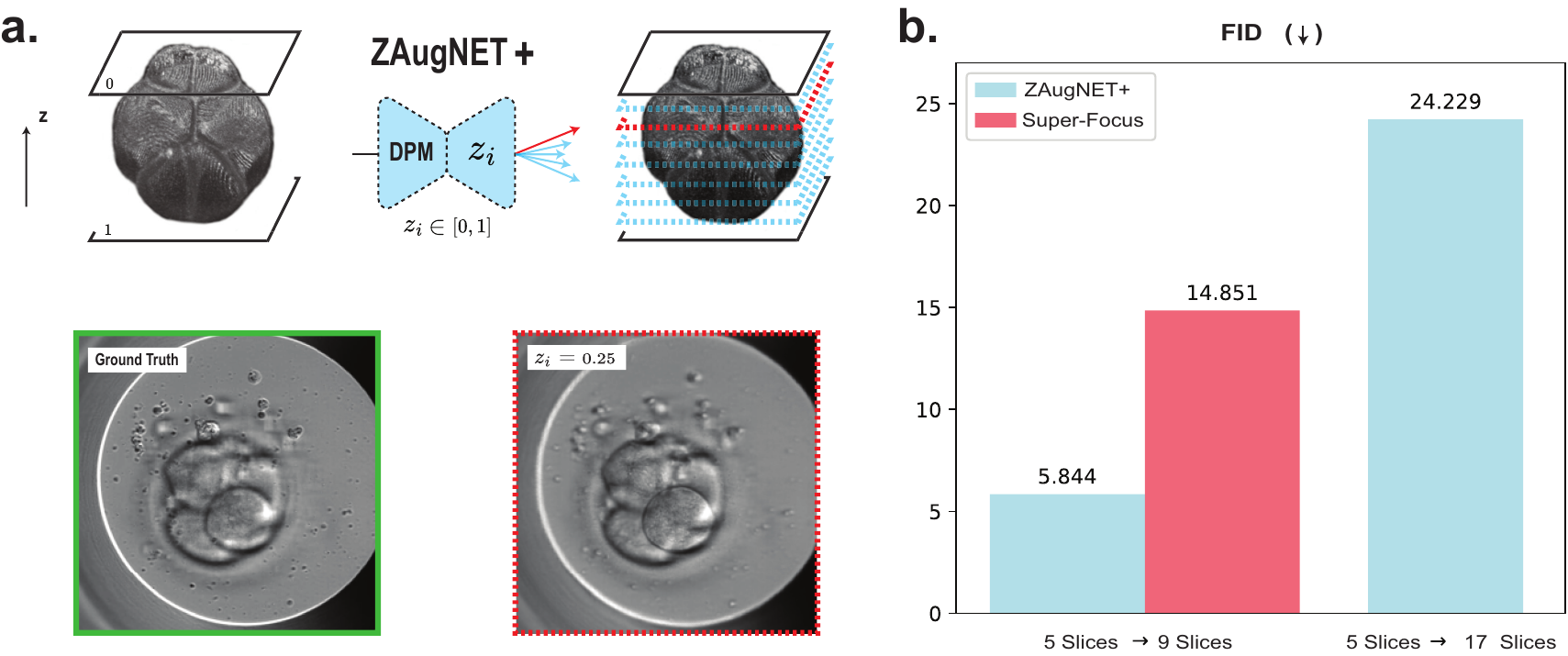}
\caption{\textbf{Continuous z-interpolation at arbitrary distances with ZAugNet+.} \textbf{a)} Example of a ZAugNet+ predicted slice at a relative distance of $ z_i = 0.25 $, shown alongside the corresponding ground truth image for visual comparison. \textbf{b)} Bar plot showing Fréchet Inception Distance (FID) values for two consecutive applications of ZAugNet+, achieving up to a 4-fold increase in axial resolution on the human embryos dataset. The red bar represents FID results reported by the authors of the Super-Focus model, being trained on a significantly larger dataset (263,024 focal stacks compared to ZAugNet’s 4,165 focal stacks). The last blue bar reflects the FID value computed by comparing the sub-resolved ground truth (9 slices) with the ZAugNet+ predicted images (17 slices), demonstrating superior interpolation quality.}\label{fig:figure5}
\end{figure}

While ZAugNet outperforms state-of-the-art techniques in terms of prediction speed and output quality, it currently interpolates z-slices at a fixed relative distance of $ z_i = 0.5 $  (i.e., midway between consecutive focal planes). To overcome this limitation, we introduce ZAugNet+, a generalized interpolation method capable of predicting z-slices at any relative distance between input focal planes (Figure \ref{fig:figure5}a). Inspired by the work of Wu et al. \cite{Wu2019} on refocusing in fluorescence imaging, this is achieved by incorporating a Digital Propagation Matrix (DPM) as an additional network input, without modifying the original architecture. The DPM is a constant matrix, shaped like the input images, where each element represents the relative distance $ z_i \in [0,1] $ from the first of the two input focal planes. A value of $ z_i = 0 $ corresponds to the first focal plane, while $ z_i = 1 $ corresponds to the second.

We trained a ZAugNet+ model on 4,165 consecutive triplets of images from a single time-lapse video capturing a human embryo at different developmental stages. For this generalized version of the algorithm, triplets were selected from the high-resolution 9-slice images, considering all possible triplet combinations $(n_1, n_2, n_3)$ satisfying $1\le n_1 < n_2 < n_3 \le 9$. This approach ensures that the model learns to interpolate slices at various relative distances between consecutive focal planes.

The trained ZAugNet+ model was first tested on 2,499 image pairs, predicting slices at the midpoint ($z_i = 0.5$). As shown in Figure \ref{fig:figure5}b, ZAugNet+ outperforms Super-Focus \cite{He2022} for human embryo slice interpolation in terms of FID values, despite the latter being trained on a significantly larger dataset (263,024 focal stacks compared to ZAugNet’s 4,165 focal stacks), indicating a far lower performance. Furthermore, ZAugNet+ demonstrates strong performance when interpolating at multiple relative distances between consecutive slices. The second blue bar in Figure \ref{fig:figure5}b illustrates the FID value obtained when ZAugNet+ simultaneously interpolates slices at $z_i = 0.25$, $z_i = 0.5$, and $z_i = 0.75$ between two consecutive slices, achieving a 4-fold axial resolution increase. The quality of the generated 17-slice stacks was assessed by comparing their distribution to that of the available sub-resolved ground truth stacks consisting of 9 slices, confirming the high fidelity of ZAugNet+ interpolations.

\section{Discussion}\label{sec3}

In this work, we introduce ZAugNet, a self-supervised deep learning-based approach designed to enhance the z-resolution of 3D biological images (Figure \ref{fig:figure1}a). By efficiently generating high-quality, high-resolution images from low-resolution stacks, ZAugNet outperforms both traditional and state-of-the-art interpolation methods (Figure \ref{fig:figure1}b).

The combination of adversarial training and knowledge distillation enables ZAugNet to improve both output quality and computational efficiency, evident in both training and prediction phases (Figure \ref{fig:figure2}a). Adversarial training preserves textural details and biological features, enhancing image realism while ensuring interpolated slices remain visually and geometrically accurate (Figure \ref{fig:figure2}b-c, Figure \ref{fig:figure4}a-c). This capability is particularly important in biological imaging, where fine-grained structural variations often carry significant biological meaning (Figure \ref{fig:figure4}b). The lightweight design of ZAugNet’s Generator, enabled by knowledge distillation, allows for efficient training on small datasets while maintaining computational efficiency during predictions. This makes ZAugNet highly scalable, even for large biological datasets (Table \ref{tab:comparison}). Across multiple image quality metrics—including RMSE, PSNR, and SSIM—ZAugNet consistently achieves low error rates, demonstrating its robustness. Compared to both bicubic interpolation and advanced deep learning-based solutions such as CAFI, ZAugNet offers superior accuracy and computational efficiency, establishing it as a state-of-the-art method (Figure \ref{fig:figure3}, Table \ref{tab:comparison}).

Despite its strong performance, ZAugNet is initially limited to interpolating slices at the midpoint between consecutive focal planes. To overcome this, we introduce ZAugNet+, a generalized version that enables continuous interpolation using a Digital Propagation Matrix (DPM). This extension provides the flexibility to predict slices at arbitrary positions, making it particularly useful for applications requiring precise control over interpolation locations. Benchmarking results confirm that ZAugNet+ not only outperforms Super-Focus—a state-of-the-art model for human embryo imaging—but also offers broader applicability across diverse imaging modalities (Figure \ref{fig:figure5}). Although ZAugNet and ZAugNet+ share identical functionality when $z_i = 0.5$, their training methodologies differ. ZAugNet+ is trained on a significantly larger dataset, utilizing numerous ordered triplets of high-resolution images. The DPM guides the model in learning the relative positioning of interpolated slices, allowing it to function as a general-purpose model capable of handling varying resolutions within a single framework. However, this added flexibility increases training time and computational demands due to the additional input channel. In contrast, ZAugNet’s specialized training allows it to optimize performance for its specific interpolation task, maintaining lower computational overhead.

While ZAugNet and ZAugNet+ demonstrate strong performance across diverse biological samples and imaging modalities, their accuracy may vary depending on the dataset and acquisition conditions. The models have not yet been evaluated across all microscopy techniques and biological systems, necessitating further studies to assess their generalizability.
Additionally, the effectiveness of ZAugNet depends on the sampling density during image acquisition. If the axial resolution in the training dataset is too sparse, fine structural details may be lost during interpolation. Similarly, significant sample motion during acquisition may introduce random correlations between slices, limiting the ability of ZAugNet and ZAugNet+ to reconstruct biologically meaningful features.
A potential improvement involves expanding the input context used for predictions. Currently, both models use only two consecutive slices as input. Incorporating additional slices from above and below the interpolation target could enhance context awareness, leading to improved accuracy and finer detail preservation in high-resolution reconstructions.

ZAugNet and ZAugNet+ mark a notable advancement in biological image interpolation, offering an optimal balance of speed, accuracy, and flexibility. Their ability to reconstruct high z-resolution images makes them valuable tools for quantitative analyses and precise cellular measurements. Beyond enhancing resolution, ZAugNet and ZAugNet+ could provide a powerful preprocessing solution that can mitigate trade-offs inherent in biological imaging. By enabling high-resolution reconstruction from lower-resolution inputs, these methods reduce the need for intense light exposure, addressing challenges such as photobleaching and phototoxicity. Additionally, ZAugNet could serve as a decompression tool, allowing large datasets to be acquired with lower storage requirements while retaining information fidelity in reconstructed images. This could be particularly beneficial for applications where data storage and transmission constraints are significant. Leveraging self-supervised deep learning, adversarial training, and knowledge distillation, ZAugNet and ZAugNet+ establish a new benchmark in biological image interpolation. Future work will focus on expanding their applicability, optimizing their performance for specialized microscopy techniques, and further refining their capabilities for real-world biological imaging challenges.

\section{Methods}\label{sec4}

\subsection{Experimental data}

\subsubsection{Ascidian embryos}
The eggs of the ascidian P. mammillata were harvested from animals obtained in Sète, France and kept in the laboratory in a tank of natural seawater at 16°C. Egg preparation and microinjection have been previously described (detailed protocols in \cite{McDougall2014, McDougall2015}). Eggs and sperm were collected by dissection. Sperm was activated in pH 9.0 seawater before fertilization (see the detailed protocol in \cite{McDougall2015}). All imaging experiments were performed at 20°C. The plasma membrane was imaged using our characterized construct PH::Tomato \cite{McDougall2015}. RNAs coding for PH::Tomato (1µg/µL) were injected in unfertilized Phallusia oocytes that were then fertilized between 2 and 12 h after injection. Four-dimensional confocal imaging was performed at 20°C using a Leica TCS SP8 inverted microscope equipped with hybrid detectors and a ×20/0.8 numerical aperture water objective lens. 3D stacks were taken with a pixel size of $1\!\times\! 1$µm and a z step of 1µm (to obtain cubic voxels).

\subsubsection{Cell nuclei}
3D image stacks of patient-derived colorectal cancer organoids without identifiable sample or patient information were provided for the analyzes of this manuscript. Organoids were embedded in Matrigel for 4-6 days. Prior to imaging organoids were fixed with 2\% paraformaldehyde, permeabilized with 0.3\% triton X-100, and stained with 1µg/mL 4',6-diamidino-2-phenylindole, dihydrochloride (DAPI; ThermoFisher Scientific) then embedded using ProLong\texttrademark~ Gold Antifade Mountant (ThermoFisher Scientific). Organoid samples were then imaged using AXR confocal microscope with a Nikon Spatial Array Confocal (NSPARC) detector using 40X CFI Plan Apochromat Lambda S 40XC Sil objective with 1 µm axial sampling.

\subsubsection{Filaments of microtubules}

Mouse Embryonic Fibroblast were seeded on a pegylated coversilp (\#1.5) containing regular spaced micropatterned disks (50µm diameter), previously covered with a protein solution (20µg/mL of fibronectin and 10µg.mL of fibrinogen conjugated with Alexa-647 in PBS). After 3 hours of spreading and reinforcement, cells were prepermeabilized using 0.025\% Triton in Cytoskeleton buffer (10 mM MES pH 6.1, 138 mM KCl, 3 mM MgCl, 2 mM EGTA, 10\% sucrose) then fixed using 4\% ParaFormAldehyde + 0.5\% Glutaraldehyde + 0.1\% Triton in Cytoskeleton buffer. Active glutaraldheydes were quenched with 1mg/ml NaBH4 prior to staining the nucleus with DAPI and the microtubule network with a rat anti-alpha tubulin antibody followed by an anti-rat conjugated with Alexa 568 secondary antibody. The sample was finally embedded in Mowiol mounting medium on a cover glass. 3D image volume (62µm$\times$62µm$\times$10µm - XYZ) of microtubules were acquired using a Zeiss LSM 900 Airyscan2 with a $63 \times$ PlanApochromat oil objective (NA 1.4) at 63 nm lateral resolution (XY pixel sizes) and 100 nm axial resolution (z-step), then finally Airyscan was post-processed. The nucleus channel was used, but not acquired, to register only single cells or cell doublets that covered all of the micropattern disks for acquisition.

\subsubsection{Human embryos}

Fertilized oocytes obtained through \textit{in vitro} fertilization techniques (either conventional or intracytoplasmic sperm injection) on day 0 were transferred to an incubation culture dish (EmbryoSlide+\texttrademark~, Vitrolife, Sweden) containing Cleav medium (CooperSurgical, Denmark) until day 3, followed by Blast medium (CooperSurgical) until day 6. Culture was maintained in oil (liquid paraffin, CooperSurgical) that had been pre-equilibrated overnight in the EmbryoScope+\texttrademark~ time-lapse incubator (Vitrolife) at 37°C, with 6\% $\text{CO}_2$, 5\% $\text{O}_2$, and 89\% $\text{N}_2$. Embryo culture was carried out under these same incubation conditions for 7 days (from day 0 to day 6). The EmbryoScope+\texttrademark~ is a time-lapse incubator that captures images of embryos every 15 minutes across 11 focal planes. The images were acquired at a resolution of 2048×1088 pixels using a 12-bit monochrome CMOS camera. This technology employs a Hoffman modulation contrast objective. After approval from the local ethics committee, the videos were retrospectively retrieved using EmbryoViewer software (Vitrolife). Each video contained all images of an embryo taken from day 0 to day 6 across the 11 focal planes.

\subsection{Network architecture and training}

ZAugNet's architecture employs a Generative Adversarial Network (GAN) \cite{goodfellow2014generativeadversarialnetworks} framework, comprising a Generator $ G $ and a Discriminator $ D $. The Generator $ G $ leverages the concept of Knowledge Distillation, utilizing a Student-Teacher scheme. A detailed overview of the architecture is provided in Figure \ref{fig:figure2}a, which is described in more detail in the following sections.

\subsubsection{Generator $G$}

Given a pair of consecutive frames $ I_0 $ and $ I_1 $, the objective is to interpolate a new frame $ \hat{I}_z $ between the two input frames with $z \in \left[0,1\right]$. ZAugNet predicts the new frame at the middle point, while in ZAugNet+ the relative position of the predicted new frame is given by the DPM matrix in input. The Student receives the two input frames and aims to estimate the intermediate flows $ F_{z \to 0} $, $ F_{z \to 1} $ , and a mask $ M $. The Student is built upon three blocks of convolutional layers followed by one transposed convolutional layer, denoted as $ S_0 $, $ S_1 $, and $ S_2 $. Each block generates an estimation of the intermediate flows and the mask, albeit with slightly different inputs. The first block, $ S_0 $, takes only the two frames $ I_0 $ and $ I_1 $ as input (together with the DPM matrix for ZAugNet+), while the blocks $ S_i $ for $ i \in \{1, 2\} $ take as input both the two input frames and the previous flows and mask produced by the previous block, as described in the following equation:

\begin{align}
    F^1_{z \to 0}, \, F^1_{z \to 1}, \, M^1 &= S_0(I_0, \, I_1) \qquad \text{ZAugNet} \nonumber \\
    F^1_{z \to 0}, \, F^1_{z \to 1}, \, M^1 &= S_0(I_0, \, I_1, \, \text{DPM}) \qquad \text{ZAugNet+} \nonumber \\
    F^{i+1}_{z \to 0}, \, F^{i+1}_{z \to 1}, \, M^{i+1} &= S_i(I_0, \, I_1, \, \hat{I}^i_{z \leftarrow 0}, \, \hat{I}^i_{z \leftarrow 1}, \, M^i) \nonumber \qquad i \in \{1, 2\}.
\end{align}
where $ \hat{I}^i_{z \leftarrow 0} $ and $ \hat{I}^i_{z \leftarrow 1} $ represent the intermediate images derived from $ I_0 $ and $ I_1 $ and the intermediate flow $ F^i_{z \to 0} $ and $ F^i_{z \to 1} $, utilizing image backward warping \cite{fischer2017flownet}, as follows:

\begin{align}
  \hat{I}^i_{z \leftarrow 0} &= \overleftarrow{\mathcal{W}} (I_0, \, F^i_{z \to 0}) \nonumber \\ 
  \hat{I}^i_{z \leftarrow 1} &= \overleftarrow{\mathcal{W}} (I_1, \, F^i_{z \to 1}) \nonumber \qquad i \in \{1,2,3\}.
\end{align}

At the end of the final block of the Student network, a weighted sum of $ \hat{I}^3_{z \leftarrow 1} $ and $ \hat{I}^3_{z \leftarrow 0} $ is computed, which represents the interpolated frame synthesized by the Student network:
\begin{align}
      \hat{I}^{S}_z = M^3 \odot \hat{I}^3_{z \leftarrow 0}  + (1-M^3) \odot  \hat{I}^3_{z \leftarrow 1}. \nonumber
\end{align}

The Teacher network consists of four blocks, with the first three blocks sharing the same architecture and weights as those of the Student network ($S_0, S_1, S_3$). The fourth block of the Teacher network ($T_3$) mirrors the Student blocks' architecture but receives an additional input, which is the ground truth frame $ I^{\text{g}}_z $:
\begin{align}
    F^4_{z \to 0}, \, F^4_{z \to 1}, \, M^4 &= T_3(I_0, \, I_1, \, \hat{I}^3_{z \leftarrow 0}, \, \hat{I}^3_{z \leftarrow 1}, \, M^3, \, I^{\text{g}}_z) \nonumber \\ 
    \hat{I}^4_{z \leftarrow 0} &= \overleftarrow{\mathcal{W}} (I_0, \, F^T_{z \to 0}) \nonumber \\ 
    \hat{I}^4_{z \leftarrow 1} &= \overleftarrow{\mathcal{W}} (I_1, \, F^T_{z \to 1}). \nonumber
\end{align}

Finally, the Teacher network generates another interpolated frame, synthesized with the additional information from the ground truth slice:
\begin{align}
    \hat{I}^{T}_z &= M^4 \odot \hat{I}^4_{z \leftarrow 0}  + (1-M^4) \odot  \hat{I}^4_{z \leftarrow 1}. \nonumber
\end{align}

\subsubsection{Discriminator $D$}
The Discriminator in ZAugNet is based on the Wasserstein GAN with Gradient Penalty (WGAN-GP) \cite{NIPS2017_892c3b1c} framework, often referred to as the "critic." The critic estimates the Wasserstein distance between the real and generated data distributions. The architecture of the discriminator network features convolutional layers followed by LeakyReLU activations.

\subsubsection{Losses}

In order to optimize the parameters of the Generator $ G $ and the Discriminator $ D $, the following loss functions for the two networks have been considered. For the Generator, the total loss is defined as:

\begin{align}
    \mathcal{L}_G = \mathcal{L}_{\text{rec}}(\hat{I}^{S}_z, I^{\text{g}}_z) + \mathcal{L}^{T}_{\text{rec}}(\hat{I}^{T}_z, I^{\text{g}}_z) + \lambda_{d} \mathcal{L}_{\text{dis}} - \lambda_{\text{adv}} \mathbb{E}_{(I_0, I_1) \sim p(I_0, I_1)} \left[ D(G(I_0, I_1)) \right], \nonumber
\end{align}
where $ \lambda_d $ and $ \lambda_{\text{adv}} $ are hyperparameters that control the balance between the different loss components, and $ \mathcal{L}_{\text{rec}} $, $ \mathcal{L}^{T}_{\text{rec}} $ are pixel-wise losses respectively for the student and the teacher, and $p(I_0, I_1)$ is the joint distribution of the input frames. For these pixel-wise losses, the Laplacian pyramid representation loss \cite{dua2018laplacian} was used. Additionally, we use the distillation loss \cite{hinton2015distilling} $ \mathcal{L}_{\text{dis}} $ to transfer the extra knowledge learned by the teacher to the student, which is then used for prediction, as defined in the following equation:

\begin{align}
    \mathcal{L}_{\text{dis}} = \sqrt{\sum_{i=1}^3  \left(\left\| F^i_{z \to 0} - F^4_{z \to 0} \right\|_2 + \left\| F^i_{z \to 1} - F^4_{z \to 1} \right\|_2\right)}. \nonumber
\end{align}

For the Discriminator, we use the WGAN-GP framework, which incorporates a gradient penalty to enforce the Lipschitz continuity constraint. A soft version of this constraint is implemented by penalizing the gradient norm on a random weighted sum of real and predicted images. The Discriminator loss is defined as:

\begin{align}
    \mathcal{L}_D = \mathbb{E}_{(I_0, I_1) \sim p(I_0, I_1)} \left[ D(G(I_0, I_1)) \right] - \mathbb{E}_{I^{\text{{g}}} \sim p_{\text{{g}}}(I^{\text{g}})} \left[ D(I^{\text{g}}) \right] + \lambda_{\text{GP}} \mathbb{E}_{\Tilde{I} \sim p(\Tilde{I})} \left[ \left( \| \nabla_{\Tilde{I}} D(\Tilde{I}) \|_2 - 1 \right)^2 \right], \nonumber
\end{align}
where $ \lambda_{\text{GP}} $ is a hyperparameter that controls the strength of the gradient penalty. We sample $ \Tilde{I} $ uniformly along a line between two samples, one from $ p_{\text{g}} $ (real data) and the other from $ p(I_0, I_1) $ (used for synthesizing data). This is done by first sampling $ I^{\text{g}} \sim p_{\text{g}}(I^{\text{g}}) $ and $ (I_0, I_1) \sim p(I_0, I_1) $, then sampling $ \alpha \sim \mathcal{U}([0, 1]) $, and computing $ \Tilde{I} $ as:

\[
\Tilde{I} = \alpha I^{\text{g}} + (1 - \alpha) G(I_0, I_1).
\]

\subsection{Training settings}

We utilized the following training protocol for ZAugNet. The model was trained from scratch over 100 epochs using 2 Nvidia V100-DGXS-32GB GPUs with a batch size of 128 (64 per GPU). Optimization was performed using the Adam algorithm \cite{kingma2017adammethodstochasticoptimization} with a learning rate of $ 1 \times 10^{-4} $ and parameters $ (\beta_1, \beta_2) = (0.5, 0.999) $. Scaling hyperparameters were set as $ \lambda_{d} = 0.01 $, $ \lambda_{\text{adv}} = 0.001 $ (for Generator-alone configuration, $\lambda_{\text{adv}} = 0$), and $ \lambda_{\text{GP}} = 10 $. Min-max normalization was applied to preprocess the data. To enhance model generalization, data augmentation techniques were incorporated, including random contrast and brightness adjustments, random rotations of up to $ 270^\circ $, and random horizontal flips. For all datasets, the input data was in 8-bit format with a resolution of $ 256 \times 256 $ pixels. For ZAugNet+ trainings, triplets were selected from high-resolution $N$-slice images by considering a limited subset of triplet combinations $(n_1, n_2, n_3)$ satisfying $\left(t-7\right) \leq n_1 < n_2 < n_3 \leq t$ with $t \in \left[ 7, N\right]$. This strategy was adopted to limit the combinatorial increase in the number of triplets within the training dataset while ensuring diverse interpolation scenarios.

\subsection{Quantification of prediction errors}

The following common metric were used to score and benchmark different interpolation methods to evaluate the quality of their predicted images : Structural Similarity (SSIM) \cite{ssim}, Root-Mean-Square-Error (RMSE), Peak-Signal-to-Noise Ratio (PSNR) \cite{psnr} and Fréchet Inception Distance (FID) \cite{fid}. 

\subsection{Roughness analysis}

Once the segmentation masks for the ZAugNet-augmented images and the corresponding ground truth images at the same resolution were obtained, a 3D surface point cloud was extracted from the 3D contours of each labeled nucleus. For each point cloud, the surface radius $ R(\theta, \phi) $, relative to the geometric center of the nucleus, was reconstructed, where $ \theta $ and $ \phi $ are the spherical polar angles. To derive a compact 3D spectral representation of the nuclear surface deformations, the real spherical harmonic coefficients $ f_{lm} $ \cite{mietke2018dynamics} were computed, defined as $ R(\theta, \phi) = \sum_{l=0}^{l_{\text{max}}} \sum_{m=-l}^{l} f_{lm} Y_{lm}(\theta, \phi) $, where $ Y_{lm}(\theta, \phi) $ is the spherical harmonic with angular number $ l $ and order $ m $. To achieve a rotation-invariant characterization of the surface roughness, the power spectrum of radial out-of-plane deformations was calculated as $ P_l = \frac{4\pi}{(2l + 1)f_{00}^2} \sum_{m=-l}^{l} f_{lm}^2 $, normalized by the average radius of the shell $ \langle R \rangle = \frac{f_{00}}{\sqrt{4\pi}} $. The non-negative quantities $ P_l $ represent the average power as a function of the angular wave number $ l $. A single-valued statistic for surface roughness, $ Ro $, was computed as $ Ro = \sum_{l \geq 3} (2l + 1)P_l $, representing the total power contained in angular numbers $ l \geq 3 $.

\bmhead{Acknowledgements}

The authors wish to honor the memory of Prof. Gabriel Popescu, whose insightful discussions in 2022 played a significant role in shaping the conceptualization of this work. His contributions to the field of bioimaging continue to inspire us. We thank all members of the Turlier team for discussions. This project received funding from the European Union’s Horizon 2020 research and innovation programme under the European Research Council grant agreement no. 949267 and under the Marie Skłodowska-Curie grant agreement No. 945304 – Cofund AI4theSciences, hosted by PSL University. H.T. has been supported by the Bettencourt-Schueller Foundation, by the CNRS and the Collège de France. A.M., R.D. and H.T. are grateful to EMBRC-France for support provided to IMEV (Investments of the Future program under reference ANR-10-INBS-0).

\section*{Declarations}

\section{Funding}
This project received funding from the European Union’s Horizon 2020 research and innovation programme under the European Research Council grant agreement no. 949267 and under the Marie Skłodowska-Curie grant agreement No. 945304 – Cofund AI4theSciences, hosted by PSL University. H.T. has been supported by the Bettencourt-Schueller Foundation, by the CNRS and the Collège de France. No ERC money was used in relation to medical human embryo microscopy data. A.M., R.D. and H.T. received funding from EMBRC-France.

\section{Competing interests}
The authors declare no competing interests.

\section{Ethics approval and consent to participate}
The use of human embryo medical images has received all necessary approvals from the French authorities. The data were anonymized before use for research purposes, and their acquisition and processing comply with the MR004 French Reference Methodology, established by the French Data Protection Authority (CNIL). Additionally, the study received approval from the Ethics and Scientific Committee of Hospices Civils de Lyon Hospital.

\section{Consent for publication}
All authors have given their consent for publication.

\section{Data availability}
To ensure accessibility, datasets and pretrained models can be accessed via Zenodo \href{https://doi.org/10.5281/zenodo.14961732}{https://doi.org/10.5281/zenodo.14961732}.

\section{Materials availability}
Correspondence and requests for materials should be addressed to Hervé Turlier: \href{mailto:herve.turlier@college-de-france.fr}{herve.turlier@college-de-france.fr}.

\section{Code availability}
To ensure accessibility, ZAugNet and ZAugNet+ are made freely available as open-source frameworks. The full codebase is hosted on GitHub \href{https://github.com/VirtualEmbryo/ZAugNet}{https://github.com/VirtualEmbryo/ZAugNet}. Additionally, an interactive Colab notebook provides a user-friendly interface for the scientific community \href{https://github.com/VirtualEmbryo/ZAugNet/blob/main/zaugnet_colab.ipynb}{https://github.com/VirtualEmbryo/ZAugNet/blob/main/zaugnet\_colab.ipynb}.

\section{Author contribution}
H.T. supervised the project and secured funding. A.P. conceived the idea, curated experimental data, and wrote the manuscript with contributions from H.T. and S.M.. A.P. and S.M. designed and implemented the algorithm, conducted testing, and benchmarked it on experimental data. A.P. performed the volumes and roughness analyses. A.P., S.M. and B.V. performed the filaments length analysis. H.T. conceptualized the manuscript and figures, which were created by A.P.. A.M., R.D., Y.A.M., B.V., A.C., and E.L. collected experimental data. All authors reviewed and approved the manuscript.

\bibliography{sn-article}

\end{document}